\documentclass[sigplan]{acmart}
\usepackage{microtype}
\usepackage{graphicx}
\usepackage{subfigure}
\usepackage{booktabs} % for professional tables
\usepackage{amsmath}
\usepackage{mathtools}
\usepackage{amssymb}
\usepackage{amsfonts}
\usepackage{listings}
\usepackage{multirow}
\usepackage{float}

\def\BibTeX{{\rm B\kern-.05em{\sc i\kern-.025em b}\kern-.08emT\kern-.1667em\lower.7ex\hbox{E}\kern-.125emX}}
    
\copyrightyear{}
\setcopyright{none}
\acmConference[HPCA.EMC2]{EMC2 '19: The 2nd workshop on Energy Efficient Machine Learning and Cognitive Computing for Embedded Applications}{Feb 17, 2019}{Washington D.C., USA}
\acmPrice{}
\acmDOI{}
\acmISBN{}

\begin{document}

%
% The "title" command has an optional parameter, allowing the author to define a "short title" to be used in page headers.
\title[Efficient Winograd or Cook-Toom Kernel for Mobile CPU]{Efficient Winograd or Cook-Toom Convolution Kernel Implementation on Widely Used Mobile CPUs}

%
% The "author" command and its associated commands are used to define the authors and their affiliations.
% Of note is the shared affiliation of the first two authors, and the "authornote" and "authornotemark" commands
% used to denote shared contribution to the research.
\author{Partha Maji}
\email{Partha.Maji@cl.cam.ac.uk}
%\email{webmaster@marysville-ohio.com}
\affiliation{%
  \institution{University of Cambridge}
%  \streetaddress{P.O. Box 1212}
%  \city{Cambridge}
%  \state{UK}
}

\author{Andrew Mundy}
\email{Andrew.Mundy@arm.com}
\affiliation{%
  \institution{Arm Research}
%  \streetaddress{1 Th{\o}rv{\"a}ld Circle}
%  \city{Cambridge}
%  \country{UK}
}

\author{Ganesh Dasika}
\email{Ganesh.Dasika@arm.com}
\affiliation{%
  \institution{Arm Research}
%  \streetaddress{1 Th{\o}rv{\"a}ld Circle}
%  \city{Austin}
%  \country{USA}
}

\author{Jesse Beu}
\email{Jesse.Beu@arm.com}
\affiliation{%
  \institution{Arm Research}
%  \streetaddress{1 Th{\o}rv{\"a}ld Circle}
%  \city{Austin}
%  \country{USA}
}
\author{Matthew Mattina}
\email{Matthew.Mattina@arm.com}
\affiliation{%
  \institution{Arm Research}
%  \streetaddress{1 Th{\o}rv{\"a}ld Circle}
%  \city{Boston}
%  \country{USA}
  }
 
\author{Robert Mullins}
\email{Robert.Mullins@cl.cam.ac.uk}
%\email{webmaster@marysville-ohio.com}
\affiliation{%
  \institution{University of Cambridge}
%  \streetaddress{P.O. Box 1212}
%  \city{Cambridge}
%  \state{UK}
}

%
% By default, the full list of authors will be used in the page headers. Often, this list is too long, and will overlap
% other information printed in the page headers. This command allows the author to define a more concise list
% of authors' names for this purpose.
\renewcommand{\shortauthors}{Maji and Mundy, et al.}

%
% The abstract is a short summary of the work to be presented in the article.
\begin{abstract}
% Deep Convolutional Neural Networks (CNNs) are known for being large and computationally expensive. It is a significant challenge to implement these models into embedded devices which usually have frugal memory and low power budget. In this paper, we explore how state-of-the-art deep CNN inference can be implemented directly on modern Arm Cortex-A CPUs, widely used in mobile devices today. Specifically, we demonstrate a reduction in compute complexity and inference time through the use of \textit{region-wise multi-channel Winograd convolution} algorithms, and by effectively leveraging the ARMv8-A NEON SIMD instruction set. We evaluated these techniques on Arm Cortex-A73 platform using several representative CNNs. The results show significant performance improvements in full network, up to 2.5${\times}$, over existing \textit{im2row/im2col} based optimisation techniques. Our techniques can be readily deployed to other Arm Cortex-A processors and provide an alternative to the use of cloud-based solutions while preserving the model architecture and pre-trained accuracy.

The Winograd or Cook-Toom class of algorithms help to reduce the overall compute complexity of many modern deep convolutional neural networks (CNNs). Although there has been a lot of research done on model and algorithmic optimization of CNN, little attention has been paid to the efficient implementation of these algorithms on embedded CPUs, which usually have very limited memory and low power budget. This paper aims to fill this gap and focuses on the efficient implementation of Winograd or Cook-Toom based convolution on modern Arm Cortex-A CPUs, widely used in mobile devices today. Specifically, we demonstrate a reduction in inference latency by using a set of optimization strategies that improve the utilization of computational resources, and by effectively leveraging the ARMv8-A NEON SIMD instruction set. We evaluated our proposed \textit{region-wise multi-channel} implementations on Arm Cortex-A73 platform using several representative CNNs. The results show significant performance improvements in full network, up to 60${\%}$, over existing \textit{im2row/im2col} based optimization techniques.
\end{abstract}



%
% Keywords. The author(s) should pick words that accurately describe the work being
% presented. Separate the keywords with commas.
\keywords{CNN, Winograd, Cook-Toom, Embedded CPU}

% %
% % A "teaser" image appears between the author and affiliation information and the body 
% % of the document, and typically spans the page. 
%   \Description{Enjoying the baseball game from the third-base seats. Ichiro Suzuki preparing to bat.}

%
% This command processes the author and affiliation and title information and builds
% the first part of the formatted document.
\maketitle

\section{Introduction}

The agility of cloud computing is great - but it simply isn't sufficient. In the near future there will be more demand for AI at the edge than in the cloud. As people need to interact with their digitally-assisted technologies (e.g. personal assistants, wearables, autonomous cars, healthcare, and other smart IoT devices) in real-time, waiting on a datacenter many miles away isn't going to work. Not only the latency matters, but often these edge devices are not within the range of the cloud needing them to operate autonomously for the most part. Even when these devices are connected to the cloud, moving high-volume of data to the centralized datacenter is not scalable, due to communication cost that impacts performance and energy consumption~\citep{DBLP:journals/corr/SzeCYE17}. Since the latency and security risk of relying on the cloud are intolerable, we need a significant portion of computation closer to the edge to permit secure, autonomous, and real-time decision making. This poses an enormous challenge in terms of implementing emerging AI workloads on resource constrained low power embedded systems. When it comes to image and video the performance of many modern embedded applications is enhanced by application of neural networks, and more specifically by convolutional neural network (CNN). Although there has been a lot of research done on algorithmic optimization of CNN~\citep{DBLP:journals/corr/SzeCYE17}, such as the Winograd, the Cook-Toom, and the Strassen, little attention has been paid to the efficient implementation of these algorithms on widely available energy efficient embedded CPUs.  This paper aims to fill this gap and investigates if emerging, compute-heavy deep CNNs can be implemented efficiently using such fast arithmetic scheme on widely used resource constrained mobile class CPUs. Specifically, we target Cortex-A class processors as Arm-based SoCs are ubiquitous in today's mobile computing~\citep{Fox:2010:TMF:2176728.2176749}.  
%\subsubsection*{Contribution} 

%We make the following contributions:

%existing implementations are limited to 2D filters only. 

%\begin{itemize}

We introduce a novel \textit{region-wise multi-channel} scheme using GEMM (General Matrix Multiplication) for energy efficient implementation of Winograd or Cook-Toom based convolution
%\footnote{In our notation, an $F(z{\times} z, w{\times} w, x{\times} x)$ algorithm convolves a $(w {\times} w)$ filter on an input of size $(x {\times} x)$ and produces an output tile of size $(z {\times} z)$.} 
on resource-constrained mobile CPUs. We show that our scheme performs better than classical \textit{im2row/col} techniques. Unlike existing implementations which are limited to 2D convolutions only, we apply variations of the base algorithms to both the 2D ($N{\times} N$) and 1D layers ($1{\times} N$, $N{\times} 1$), where $N$ is the height/width of the filter. We demonstrate the efficiency of our scheme by implementing a number of widely used state-of-the-art deep CNNs on the energy-efficient Arm Cortex-A73 processor~\citep{TheLinle45:online}. 
%To the best of our knowledge, this paper is the first detailed study of Winograd-based acceleration of deep CNN inference on widely used mobile Arm processors.

Our results show that by effectively using Armv8-A NEON SIMD instructions and appropriate choice of variations of Winograd or Cook-Toom based convolution an average $2-3{\times}$ and a peak $4{\times} $ per layer speedup on top of aggressively optimized solutions using the classical \textit{im2row/col} technique is achievable.  
%Our results show that by effectively using Armv8-A NEON SIMD instructions and region-wise multi-channel GEMM kernels an average $2-3{\times}$ and a peak $3.9{\times} $ per layer speedup on top of aggressively optimized solutions using the classical \textit{im2row/col} technique is achievable. 
As an example, our multithreaded implementation of SqueezeNet on Arm Cortex-A73 can achieve an average inference rate of 47 frames/sec -- sufficient for many real-time embedded applications~\citep{8057318}. Our scheme can be readily deployed to other widely used ARMv8-A cores. 

\section{Strategies for Efficient Multichannel Winograd or Cook-Toom Kernel Implementation on Armv8-A Cores}
The Winograd or Cook-Toom class of algorithms~\cite{DBLP:journals/corr/Lavin15b, e20040305} help to reduce the overall compute complexity of convolution by reducing the number of required multiplication. 
%This helps to reduce the total number of multiplications required to compute each planar convolution. 
Implementations of these algorithms are well suited to CNNs consisting of small filters and low power embedded systems as the resources and power budget are very limited. Using the Winograd or Cook-Toom based convolution, a typical layer of a convolutional neural network (CNN) can be expressed in the following matrix equation

\begin{equation}\label{eq:zwxc2}
f =  Z^T(\sum^{C}_{c=0}[(WwW^T)_c \odot (X^TxX)_c])Z
\end{equation}
where $W$ and $X$ are the transform matrices for the weight and the input sequence $w$, $x$, respectively. $Z$ is the inverse transformation matrix, and $\odot$ is the elementwise (Hadamard) product.
%, a binary operation that takes two matrices of the same dimensions, and produces another matrix where each element $i,j$ is the product of elements $i,j$ of the original two matrices. 

%The previous section introduced the Winograd transform and demonstrated how multi-channel computation could be expressed using addition within the Winograd domain.
%In this section we combine the techniques developed above to build an efficient Winograd convolution implementation for Arm CPU.

First, we note that the equation shown above applied a $(w{\times} w)$ filter to only a small input region of size $(x{\times} x)$ to produce an output region of size $(z{\times} z)$ (a.k.a. $F(z{\times} z, w{\times} w, x{\times} x)$). 
To perform larger convolutions we must, therefore, break the input tensor into multiple regions of size $(x {\times} x)$.
The output tensor must also be divided into an equivalent number of regions, each of which is computed as the elementwise multiplication and accumulation of the corresponding input regions (representing $C$ input channels) with their respective weight tiles. This algorithm is illustrated in Listing~\ref{sample-winograd-loop}.
\begin{lstlisting}[caption=Sample Winograd convolution algorithm, 
label=sample-winograd-loop, float=!ht, belowskip=-0.8 \baselineskip]
// For each output channel
for (unsigned int m = 0; m < M; m++)
  // For each output region
  for (unsigned int r = 0; r < R; r++)
    // Summation across the input channels
    for (unsigned int c = 0; c < C; c++)
      output_region[m, r] += HadamardProduct(
        input_region[c, r], weight_region[m,c]);
\end{lstlisting}

% For example, a $6 {\times} 6$ input tensor would be broken into a $2{\times} 2$ array of $4 {\times} 4$ regions.
% Separate arrays of regions are produced for each channel in a tensor.
% Hence, a $3 {\times} 6 {\times} 6$ tensor would produce a $3 {\times} 2 {\times} 2$ array of regions.
% Convolving this input tensor with $M$ filters would result in the construction of $M {\times} 2 {\times} 2$ array of regions.
% Output regions are computed as described above:
% for example, the upper-leftmost output region in the $m^\mathrm{th}$ channel results from the summation of $C$ products -- each of the $C$ channels of upper-leftmost input region with the respective tiles from the $m^\mathrm{th}$ filter.
% This algorithm is illustrated in Listing~\ref{sample-winograd-loop}.

We break our region wise multi-channel algorithm into three steps:

\begin{enumerate}
 \item \textbf{Input Transform} Progresses over regions of the input tensor, transforms them into the Winograd domain and \emph{scatters} the results into the `A' matrices for the GEMMs.
 \item \textbf{GEMM} \emph{Multiplies} the `A' matrices generated in the \emph{input transform} with `B' (matrices generated when the weights were transformed into the Winograd domain) to form the `C' matrices.
 \item \textbf{Output Transform} Repeatedly \emph{gathers} regions of values from the `C' matrices, transforms them back into the spatial domain and writes the results into the output tensor.
\end{enumerate}

%The first and last of these steps are discussed below.

\subsection{Data Layout and SIMD Computation}

There are a variety of ways in which 4D tensors can be arranged in memory.
Two common options are called \emph{NCHW} and \emph{NHWC} -- where \emph{N} stands for the number of batches (or concurrent inferences), \emph{C} for the number of channels, and \emph{H} and \emph{W} stand for height and width, respectively. In \emph{NCHW} each plane of the tensor is stored contiguously in memory -- i.e., pixel $(n, c, i, j)$ is followed by $(n, c, i, j + 1)$ -- whereas in \emph{NHWC} all of the channels of a given pixel are stored contiguously (i.e., value $(n, i, j, c)$ is followed by $(n, i, j, c + 1)$).
%For scalar code, ignoring cache locality, the choice of tensor ordering is largely immaterial.
When writing vectorized (SIMD) code, tensor ordering is crucial to achieving performance.

\begin{figure}[!htbp]\hspace{1.0cm}
	\includegraphics[trim={0.0cm 0.0cm 0.0cm 0cm},clip,scale=0.6]{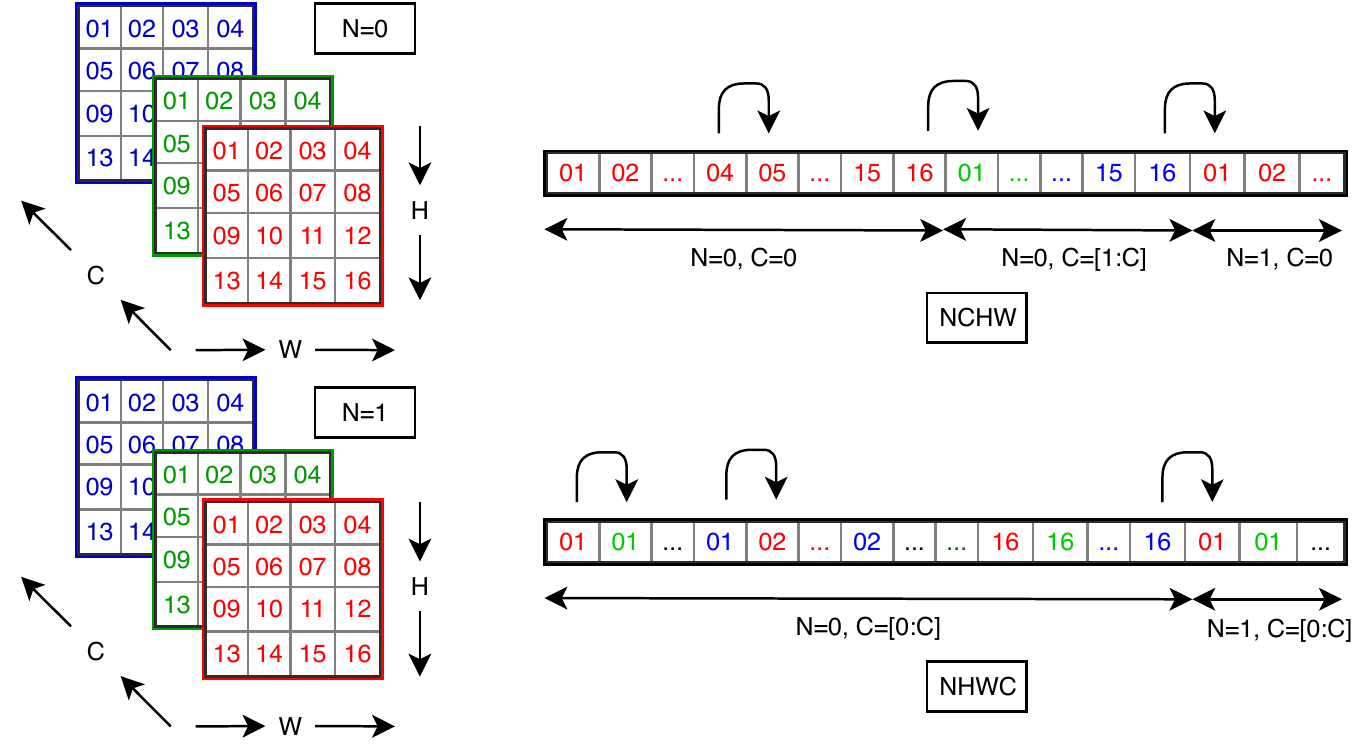}
    \centering
	\caption{NCHW vs NHWC Layout}
   	\label{fig:nchw}
\end{figure}

In the Armv8-A architecture, there are thirty-two 128-bit SIMD registers.
Each SIMD register can, therefore, store four 32-bit single-precision values.
Hence, under \emph{NCHW} a single SIMD register will store, after a 128-bit load, a row of four pixels, whereas under \emph{NHWC} the same register would store four channels of data for a single pixel.
We can see the effect of these orderings through the example of implementing the input transform for $F(2{\times} 2, 3{\times} 3, 4{\times} 4)$.

\subsubsection{Input Transform For $F(2{\times} 2, 3{\times} 3, 4{\times} 4)$}
%\textbf{\textit{Input transform for $F(2{\times} 2, 3{\times} 3, 4{\times} 4)$.}}

The characteristic equation for this transform is:

\begin{equation}
  X^\top x X = \begin{bmatrix*}[r]
    1 & 0 & -1 & 0 \\
    0 & 1 &  1 & 0 \\
    0 & -1 & 1 & 0 \\
    0 & 1 & 0 & -1
  \end{bmatrix*} x \begin{bmatrix*}[r]
    1 & 0 & 0 & 0 \\
    0 & 1 & -1 & 1 \\
    -1 & 1 & 1 & 0 \\
    0 & 0 & 0 & -1
  \end{bmatrix*}
\end{equation}

Under \emph{NCHW} ordering we would use four registers to store $x$, a $4 {\times} 4$ region of the input tensor.
The transform matrices could be hard-coded as a series of row-transformations, such that computing $X^\top x$ could be expressed as:

\begin{lstlisting}[float=h]
 XTx[0] = vsubq_f32(x[0], x[2]); // x_1i - x_3i
 XTx[1] = vaddq_f32(x[1], x[2]); // x_2i + x_3i
 XTx[2] = vsubq_f32(x[2], x[1]); // x_3i - x_2i
 XTx[3] = vsubq_f32(x[1], x[3]); // x_2i - x_4i
\end{lstlisting}

By transposing the result, this code sequence can be repeated such that we compute $\left(X^\top \left((X^\top x)^\top\right) \right)^\top = X^\top x X$.
Once this is completed we have 16 values (four registers containing four values each) which must be scattered, as described before, to 16 separate matrices.

In contrast, under \emph{NHWC} ordering, we would use sixteen SIMD registers to represent four channels of a $4 {\times} 4$ region of the input tensor.
The transformation can be hardcoded, but in this case we operate on four channels of data simultaneously, as in Listing~\ref{xt}.
Once the transformation is complete we are left with sixteen registers, each containing four channels worth of data.
These registers can be scattered directly into the the input matrices for the GEMMs.

\begin{lstlisting}[caption=Input Transforms., label=xt, float=!bhtp, belowskip=-0.8 \baselineskip]
 // Compute X^T x and U = (X^T x) X
 for (int j = 0; j < 4; j++) {  
 // For each column in X^T x
    XTx[0][j] = vsubq_f32(x[0][j], x[2][j]);
    XTx[1][j] = vaddq_f32(x[1][j], x[2][j]);
    XTx[2][j] = vsubq_f32(x[2][j], x[1][j]);
    XTx[3][j] = vsubq_f32(x[1][j], x[3][j]); 
    }
 for (int i = 0; i < 4; i++) { 
 // For each row in U
    U[i][0] = vsubq_f32(XTx[i][0], XTx[i][2]);
    U[i][1] = vaddq_f32(XTx[i][1], XTx[i][2]);
    U[i][2] = vsubq_f32(XTx[i][2], XTx[i][1]);
    U[i][3] = vsubq_f32(XTx[i][1], XTx[i][3]); 
    }
\end{lstlisting}

\subsubsection{Choice of NHWC over NCHW}

For the specific instance of $F(2{\times} 2, 3{\times} 3, 4{\times} 4)$ there are merits to both approaches.
However, when we consider using either different data widths (such as half-precision floating point) or different version of Winograd or Cook-Toom algorithms we begin to see advantages to the \emph{NHWC} ordering.

For example, although we can use four SIMD registers to represent 16 values \emph{in single-precision floating point} in \emph{NCHW} -- four values to a register -- this breaks down when we move to half-precision and each register can contain eight values, 
%.
%The code sample from above for \emph{NCHW} would need to be extensively rewritten to deal with this case
whereas the \emph{NHWC} code could be simply modified to work on eight channels of data simultaneously.

Likewise, were we to implement the input transform for $F(4{\times} 4, 3{\times} 3, 6{\times} 6)$, which requires use of $6{\times} 6$ input regions, we could, in \emph{NHWC} ordering use 36 values (and the stack) to represent each input region.
However, in \emph{NCHW}, we would need to use one-and-a-half registers to represent each row of six values.
%Again, the code would be considerably more complicated. 
For these reasons, we prefer the use of \emph{NHWC} ordered data.

% \begin{figure*}
% 	\includegraphics[trim={0.1cm 0.3cm 0.1cm 0.1cm},clip,scale=0.477]{inception-v3-3x3-L}
%     \centering
% 	\caption{Layerwise speedup in the two dimensional convolution layers with 3x3 and 5x5 filters in \textbf{Inception-v3} for batch size of 1}
%    	\label{fig:inception-v3-layerwise-speedup-3x3}
% \end{figure*}

\subsubsection{Efficient Tensor Ordering for ARMv8-A Cores}
The convolution of a tensor consisting of $C$ input layers and $R$ regions with a $M$ deep set of filters can be expressed as $x^2$ GEMMs of the form $[R {\times} C] {\times} [C {\times} M]$ or $[M {\times} C] {\times} [C {\times} R]$, and that, of these, we preferred the former as shown in Figure~\ref{fig:winograd-gemm}.
This selection follows directly from our choice of \emph{NHWC} tensor ordering.
Specifically, we note that, under \emph{NHWC}, each SIMD register contains multiple channels of data and that these values must be written into matrices of shape $R {\times} C$ or $C {\times} R$.
Assuming row-major ordered matrices we note that, in the latter case, we could use multi-element structured stores (e.g., \texttt{ST4} (single structure), \cite{ArmARM}) to combine and store values from different registers.
Alternatively, an unstructured store (\texttt{STR} \cite{ArmARM}) could be used to write out a whole register into successive columns of an $R {\times} C$ matrix.
Since we found unstructured stores to have a higher throughput than their structured counterparts we choose to use the first form.

\subsection{Using GEMM to Compute Hadamard Products}

\begin{figure*}
	\includegraphics[scale=0.7]{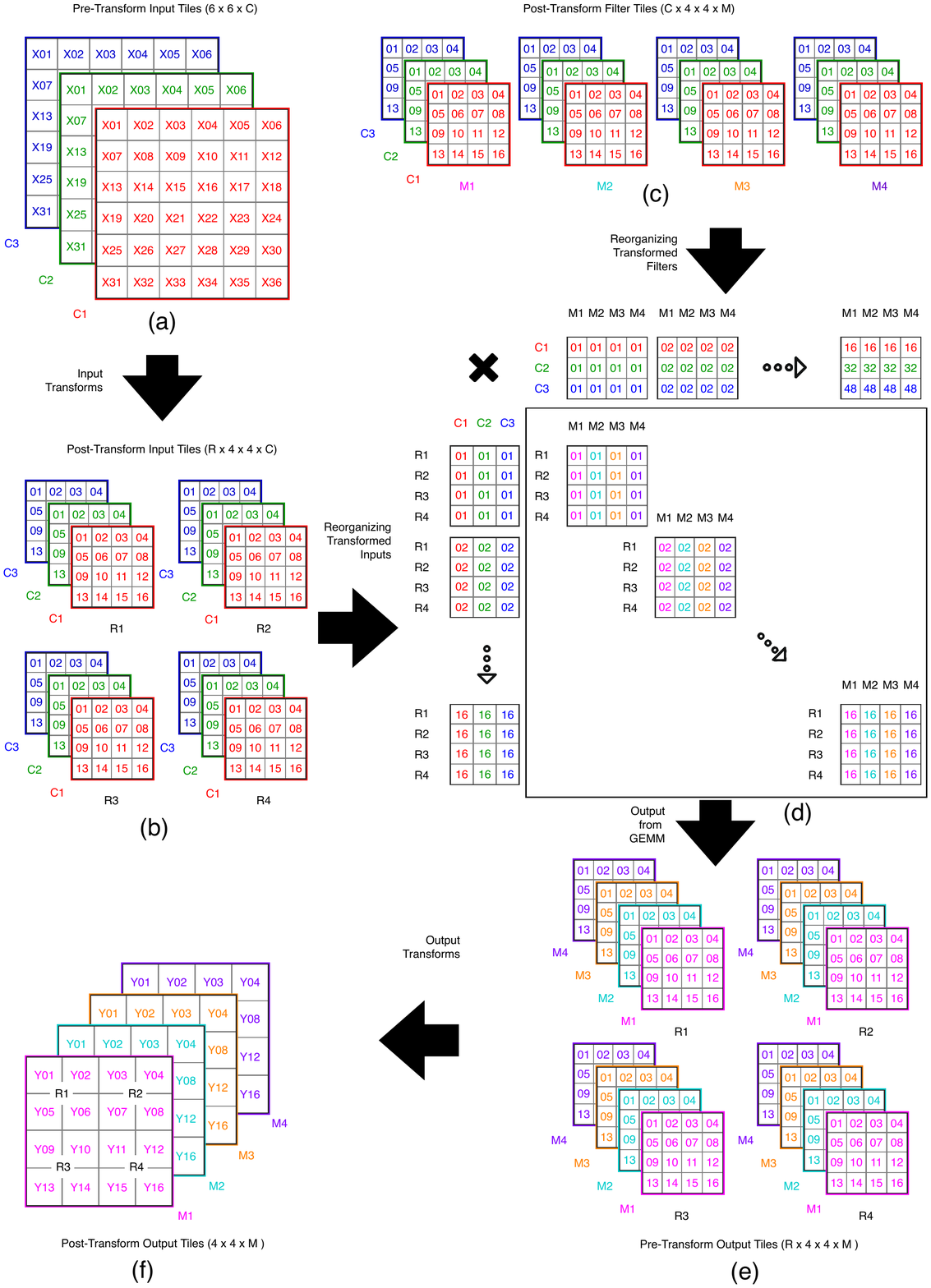}
    \centering
	\caption{Detailed Data-Flow Diagram in Region-wise Multi-channel Winograd or Cook-Toom based Scheme -- (a) Pre-transform Input Channels, (b) Transformed Input Channel Regions, (c) Transformed Filters, (d) GEMM Kernels, (e) Output of GEMM in the Residue Domain, (f) Final Output Channels after applying Inverse Transforms}
   	\label{fig:winograd-gemm}
\end{figure*}

By inspecting the basic convolution algorithm illustrated in Listing~\ref{sample-winograd-loop} we observe, firstly, that the fundamental operation is an element-wise multiply-accumulate (element-wise addition of Hadamard products).
Secondly, we note that there are two axes in which data is reused - (1) Weight tile $(m, c)$ is used across all input regions in layer $c$, and, (2) Input region $(c, i, j)$ contributes to all $M$ output regions at $(i, j)$.
These observations suggest that one way of implementing a complete convolution is to leverage the GEMM (General Matrix Matrix Multiplication) algorithm since there exist a wide range of good GEMM implementations (e.g.,~\cite{DBLP:journals/corr/FrisonKZD17}) capable of exploiting the SIMD instructions of the Armv8-A architecture. Figure~\ref{fig:winograd-gemm} shows an example for a $3 {\times} 6 {\times} 6$ tensor being convolved with four filters. An array of 16 GEMMs of size $[R {\times} C] {\times} [C {\times} M]$ is constructed, with the input tensor being represented by the first set of matrices and the weights by the latter.
%We now describe how the algorithm illustrated in Listing~\ref{sample-winograd-loop} can be mapped to GEMM.

% Firstly, we note that, since the key data processing operation is \emph{element-wise} multiply-accumulate, each cell of input, weight and output regions is treated independently of the others.
% This suggests that for a Winograd algorithm with a $(x {\times} x)$ input region size we must perform $x^2$ independent GEMMs.

% Next, we note that each output cell is the result of $C$ multiply-accumulates -- this suggests that inner dimension of our matrix multiplication must be $C$ (where $C$ is the number of channels in the input tensor).

% Finally, we note that, excepting the number of channels, the input and output region arrays are the same size -- this suggests that one dimension of each GEMM should be the number of regions in a layer, $R$, and the final dimension of the GEMM the number of output channels, $M$.
% Consequently, the convolution of a tensor consisting of $C$ input layers and $R$ regions with a $M$ deep set of filters can be expressed as $x^2$ GEMMs of the form $[R {\times} C] {\times} [C {\times} M]$ or $[M {\times} C] {\times} [C {\times} R]$.
% For reasons which will be explained below, we choose the former.
% In this form the first matrix is a representation of the input tensor and must be computed for each inference and the latter represents the weights and can be computed once and reused.
% Figure~\ref{fig:winograd-gemm} illustrates this mapping.

%%%%%%%%%%%%%%%%%%%%%%%%%%%%%%%%%%%%%%%%%%%%%%%%%%%%%%%%%%%%%%%%%%%%%%%%%%%%%%%

\begin{figure*}[!htbp]\hspace{1.0cm}
	\includegraphics[trim={0.5cm 7cm 0.1cm 7cm},clip,scale=0.53]{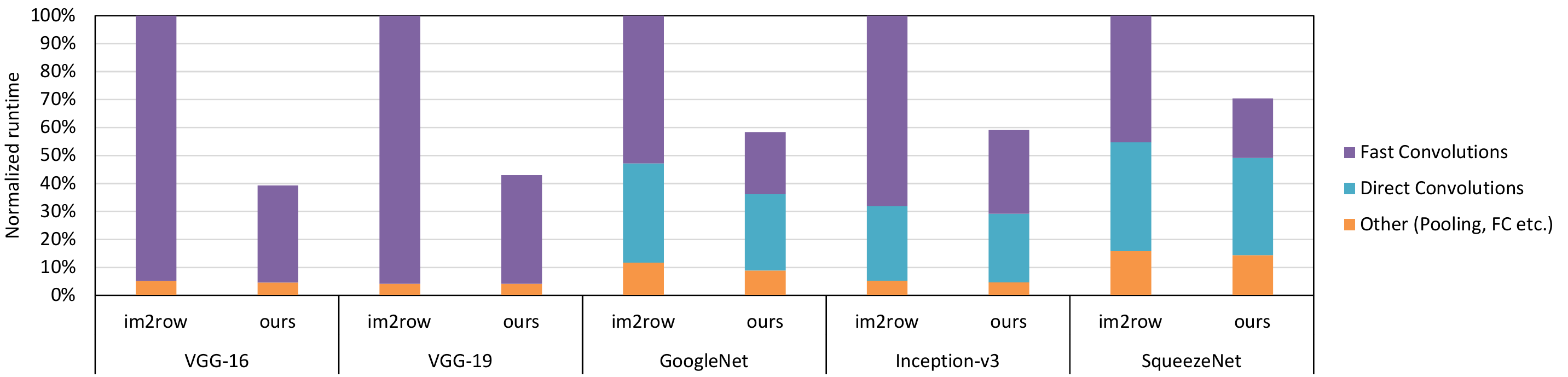}
    \centering
	\caption{Speed-up achieved in the Winograd or Cook-Toom suitable layers as a \textbf{fraction} of the entire model (batch size = 1)}
   	\label{fig:full-cnn-speedup}
\end{figure*}
%%%%%%%%%%%%%%%%%%%%%%%%%%%%%%%%%%%%%%%%%%%%%%%%%%%%%%%%%%%%%%%%%%%%%%%%%%%%%%%
%%%%%%%%%%%%%%%%%%%%%%%%%%%%%%%%%%%%%%%%%%%%%%%%%%%%%%%%%%%%%%%%%%%%%%%%%%%%%%%

\begin{table*}[!ht]
  \centering
  \caption{Summary of \textbf{mean absolute runtime} of the \textbf{whole-network} in milliseconds (msec) for batch size of 1}
  \label{tab:absolute-result}
  \resizebox{\linewidth}{!}{%
  \begin{tabular}{l||rr|rr|rr|rr}
    \toprule
    \multirow{2}{*}{} &
      \multicolumn{2}{c}{VGG-16} &
      \multicolumn{2}{c}{GoogleNet} &
      \multicolumn{2}{c}{Inception-v3} &
      \multicolumn{2}{c}{SqueezeNet} \\
%    & {Full Network} & {Winograd Layers} & {Full Network} & {Winograd Layers} & {Full Network} & {Winograd Layers} & {Full Network} & {Winograd Layers} \\
        & {Full Network} & {Fast Layers} & {Full Network} & {Fast Layers} & {Full Network} & {Fast Layers} & {Full Network} & {Fast Layers} \\
%           & \multicolumn{1}{c}{Full} & \multicolumn{1}{c}{Winograd} & \multicolumn{1}{c}{Full} & \multicolumn{1}{c}{Winograd} & \multicolumn{1}{c}{Full} & \multicolumn{1}{c}{Winograd} & \multicolumn{1}{c}{Full} & \multicolumn{1}{c}{Winograd} \\
%           & \multicolumn{1}{c}{Network} & \multicolumn{1}{c}{Layers} & \multicolumn{1}{c}{Network} & \multicolumn{1}{c}{Layers} & \multicolumn{1}{c}{Network} & \multicolumn{1}{c}{Layers} & \multicolumn{1}{c}{Network} & \multicolumn{1}{c}{Layers} \\
    \midrule
    Using Im2Row Scheme & 1929.43 & 1829.10 & 173.13 & 91.42 & 750.37 & 510.92 & 29.72 & 13.47  \\
    Using Our Scheme & 758.05 & 670.79 & 101.04 & 38.38 & 443.40 & 224.42 & 20.91 & 6.29  \\
%    Frames/Sec & 1 & - & 9 & - & 2 & - & 47 & -  \\
    Speedup (msec) & 1171.38 & 1158.31 & 72.09 & 53.04 & 306.98 & 286.51 & 8.81 & 7.18  \\
    Speedup (\%) & 60.71\% & 63.33\% & 41.64\% & 58.02\% & 40.91\% & 56.08\% & 29.64\% & 53.28\% \\
    \bottomrule
\end{tabular}%
}
\end{table*}
%%%%%%%%%%%%%%%%%%%%%%%%%%%%%%%%%%%%%%%%%%%%%%%%%%%%%%%%%%%%%%%%%%%%%%%%%%%%%%%

% Shown in Figure~\ref{fig:winograd-gemm}(a) is the example: a $3 {\times} 6 {\times} 6$ tensor being convolved with four filters.
% In (b) this tensor is split into a $3 {\times} 2 {\times} 2$ array of regions, each of which is transformed into the Winograd domain.
% Meanwhile, in (c), a set of four filters (each, necessarily, of three channels) is shown transformed into the Winograd domain.
% In (d), these weights are combined with the regions from (b) to form the GEMM structure described above.
% Specifically, an array of 16 GEMMs of size $[R {\times} C] {\times} [C {\times} M]$ is constructed,
% with the input tensor being represented by the first set of matrices and the weights by the latter.
% Performing the GEMMs results in the creation of 16 $[R {\times} M]$ matrices which are reordered, in (e), into regions of output in the Winograd domain.
% In (f) these regions are converted back to the spatial domain and so arranged as to form the result of the convolution.

%%%%%%%%%%%%%%%%%%%%%%%%%%%%%%%%%%%%%%%%%%%%%%%%%%%%%%%%%%%%%%%%%%%%%%%%%%%%%%%

\begin{table}[!ht]
  \centering
  \caption{per-layer speedup comparison: im2row vs ours}
  \label{tab:per-player-result}
  \resizebox{0.85\linewidth}{!}{%
  \begin{tabular}{l||rrr}
    \toprule
    \multirow{2}{*}{Model} &
      \multicolumn{3}{c}{Per-layer Speedup} \\
        & {Layer-type} & {Average Speedup} & {Peak Speedup} \\
    \midrule
    VGG-16 & $3\times3$ & $2.7\times$ & $3.5\times$  \\
    VGG-19 & $3\times3$ & $2.8\times$ & $3.5\times$ \\
    GoogleNet & $3\times3$ & $2.6\times$ & $4.1\times$ \\
    GoogleNet & $5\times5$ & $2.3\times$ & $3.2\times$ \\
    Inception-v3 & $1\times7$  & $2.0\times$ & $2.1\times$ \\
    Inception-v3 & $7\times1$  & $2.0\times$ & $2.1\times$ \\
    Inception-v3 & $3\times3$  & $3.1\times$ & $3.8\times$ \\
    Inception-v3 & $5\times5$  & $2.7\times$ & $2.8\times$ \\
    SqueezeNet & $3\times3$ & $2.2\times$ & $2.6\times$ \\
    \bottomrule
\end{tabular}%
}
\end{table}
%%%%%%%%%%%%%%%%%%%%%%%%%%%%%%%%%%%%%%%%%%%%%%%%%%%%%%%%%%%%%%%%%%%%%%%%%%%%%%%

\section{Evaluation and Results}

%\subsection{Benchmark Platform and Models}
%\subsection{Evaluation Setup - models and platform}
We chose five widely used CNNs of different sizes and complexities to validate our implementation, namely, VGG19, VGG16, GoogleNet, Inception-v3, and SqueezeNet~\cite{DBLP:journals/corr/SzeCYE17}. 
We benchmarked our implementation on the Huawei HiKey 960 development platform using IEEE 754 fp32 standard.

\subsection{Results -- per-layer speedup}

% We implemented five different variants of the fast algorithm -   
% $F(4{{\times}} 4, 3{{\times}} 3, 6{\times} 6)$, 
% $F(2{\times} 2, 5{\times} 5, 6{\times} 6)$ for 2D convolutions and  
% $F(1{\times} 6, 1{\times} 3, 1{\times} 8)$,
% $F(1{\times} 4, 1{\times} 5, 1{\times} 8)$, and, $F(1{\times} 2, 1{\times} 7, 1{\times} 8)$ for 1D convolutions. To investigate their performance 
We implemented five different variants of the fast algorithm and bench-marked them on individual layers of all the selected models.
In each case we measured the number of cycles taken to perform all three stages of our algorithm (Input transform, GEMMs and Output transform) on the 'big'-cluster which consists of four Cortex-A73 core.
%Partha: We need to mention multi-threaded somewhere here.
As a baseline against which to compare we also benchmarked the GEMM calls which would result from application of the classical \textit{im2row} technique to the same layers.
Table~\ref{tab:per-player-result} presents the speedup achieved by our region-wise multi-channel Winograd scheme over the GEMM.

\subsection{Results -- whole-network speedup}
To measure the effectiveness of Winograd or Cook-Toom based acceleration for end-to-end CNN, we also benchmarked the runtime of entire models.
In these cases we used the Arm Compute Library~\cite{Technolo3:online} to evaluate single-batch (batch size of 1) inferences of these networks on multi-threaded ($4{\times}$) Cortex-A73.
Two sets of benchmarks were run: in one, layers suitable for the Winograd-based acceleration use our scheme, and the rest use baseline \textit{im2row} scheme; in the other all layers use \textit{im2row}.
%Since whole-network performance is a combination of the time spent performing convolution and that spent computing other functions (such as pooling and activation) we expect to see lower speedups than those benchmarked for individual layers.
Figure~\ref{fig:full-cnn-speedup} and Table~\ref{tab:absolute-result} shows the normalized  and the absolute runtime of the five CNNs (whole-network), respectively. 
% In each case the entire network was run 21 times and we show the mean, noise in the system contributes to slight variations in the time spent in each portion of the network. For VGG-16 and VGG-19, which are large and dominated by Winograd-suitable $3{\times} 3$ and $5{\times} 5$ convolution layers, we see whole network speedups approaching 2.5${\times}$. For Inception-v3 and GoogleNet which are moderately medium size models achieve a speedup of 1.7${\times}$. For SqueezeNet which is an extremely compact model achieves a speedup of 1.4${\times}$. 
%In terms of absolute inference speed with batch size of one, SqueezeNet and GoogleNet achieve an impressive 47 frames/sec and 10 frames/sec, respectively. The larger VGG-16 and Inception-v3 models achieve between 1 and 3 frames/sec. See Table~\ref{tab:absolute-result} for more details. 

% It can be deduced from the table, that SqueezeNet and GoogleNet achieve an impressive 47 frames/sec and 10 frames/sec, respectively. The larger VGG-16 and Inception-v3 models achieve between 1 and 3 frames/sec. In all cases, the pre-trained accuracy is preserved. 

\section{Conclusions}

Winograd or Cook-Toom based acceleration on Arm's Cortex A CPUs can dramatically reduce the compute time and energy cost of individual convolution layers -- by up to 4${\times}$.
However, these speedup numbers are lower than the theoretical values. Partially, this is due to the challenges involved in implementing the algorithm in a real system but largely it is because the theoretical speed-up of this class of algorithm disregards the cost of transforming to and from the alternative domain of computation. 
% The theoretical speed-up of an $F(z {\times} z, w {\times} w, x {\times} x)$ Winograd algorithm can be computed as $(zw)^2/(z+w-1)^2$.
% As an example, the theoretical speed-up for an $F(4 {\times} 4, 3 {\times} 3, 6 {\times} 6)$ variation of the algorithm is $4{\times}$, whereas we have only benchmarked a $3.5{\times}$ speed-up on a single-threaded implementation. 
This gap between the theoretical and achieved speed-ups can be somewhat overcome by amortizing the transform costs over those of the GEMMs.
%For example, since each input region is used to compute multiple output regions (as many as there are output channels) the cost of the input transformation is amortized over the number of output channels.
As the number of output channels increases, the speed-up will asymptotically approach the maximum achievable.

\bibliographystyle{ACM-Reference-Format}
\bibliography{sample-base}

% 
% If your work has an appendix, this is the place to put it.
% \appendix



% Lorem ipsum dolor sit amet, consectetur adipiscing elit. Morbi malesuada, quam in pulvinar varius, metus nunc fermentum urna, id sollicitudin purus odio sit amet enim. Aliquam ullamcorper eu ipsum vel mollis. Curabitur quis dictum nisl. Phasellus vel semper risus, et lacinia dolor. Integer ultricies commodo sem nec semper. 


% Etiam commodo feugiat nisl pulvinar pellentesque. Etiam auctor sodales ligula, non varius nibh pulvinar semper. Suspendisse nec lectus non ipsum convallis congue hendrerit vitae sapien. Donec at laoreet eros. Vivamus non purus placerat, scelerisque diam eu, cursus ante. Etiam aliquam tortor auctor efficitur mattis. 


% Nam id fermentum dui. Suspendisse sagittis tortor a nulla mollis, in pulvinar ex pretium. Sed interdum orci quis metus euismod, et sagittis enim maximus. Vestibulum gravida massa ut felis suscipit congue. Quisque mattis elit a risus ultrices commodo venenatis eget dui. Etiam sagittis eleifend elementum. 

% Nam interdum magna at lectus dignissim, ac dignissim lorem rhoncus. Maecenas eu arcu ac neque placerat aliquam. Nunc pulvinar massa et mattis lacinia.

\end{document}